\documentclass[letterpaper, 10 pt, conference]{ieeeconf}  

\IEEEoverridecommandlockouts
\usepackage{graphics} 
\usepackage{epsfig} 
\usepackage{mathptmx} 
\usepackage{times} 
\usepackage{amsmath} 
\usepackage{amssymb}  
\usepackage{color}  
\usepackage[dvipsnames]{xcolor}
\usepackage{balance}
\usepackage{cite}
\usepackage{hyperref}

\usepackage[normalem]{ulem}

\title{\LARGE \bf
LMNet: Real-time Multiclass Object Detection
\\  on CPU using 3D LiDAR}

\author{Kazuki Minemura$^{1}$, Hengfui Liau$^{1}$, Abraham Monrroy$^{2}$ and Shinpei Kato$^{3}$\\
\textit{\{kazuki.minemura, heng.hui.liau\}@intel.com,}
\textit{amonrroy@ertl.jp,}
\textit{shinpei@is.s.u-tokyo.ac.jp}
\thanks{$^{1}$ Kazuki Minemura and Hengfui Liau are with Intel's Internet of Things Group, Bayan Lepas, 11900, Penang, Malaysia.}
\thanks{$^{2}$ Abraham Monrroy is with Nagoya University, Parallel and Distributed Systems lab, Nagoya, Furo-cho, 464-8601, Aichi, Japan.}
\thanks{$^{3}$ Shinpei Kato is Associate Professor at the School of Science for The University of Tokyo, Bunkyo-ku, 113-0033, Tokyo, Japan.}
}

\begin{document}

\maketitle

\thispagestyle{empty}
\pagestyle{empty}

\begin{abstract}
This paper describes an optimized single-stage deep convolutional neural network to detect objects in urban environments, using nothing more than point cloud data. This feature enables our method to work regardless the time of the day and the lighting conditions.
The proposed network structure employs dilated convolutions to gradually increase the perceptive field as depth increases, this helps to reduce the computation time by about 30\%. The network input consists of five perspective representations of the unorganized point cloud data. The network outputs an objectness map and the bounding box offset values for each point. Our experiments showed that using reflection, range, and the position on each of the three axes helped to improve the location and orientation of the output bounding box. We carried out quantitative evaluations with the help of the KITTI dataset evaluation server. It achieved the fastest processing speed among the other contenders, making it suitable for real-time applications. We implemented and tested it on a real vehicle with a Velodyne HDL-64 mounted on top of it. We achieved execution times as fast as 50 FPS using desktop GPUs, and up to 10 FPS on a single Intel Core i5 CPU. The deploy implementation is open-sourced and it can be found as a feature branch inside the autonomous driving framework Autoware. Code is available at: \url{https://github.com/CPFL/Autoware/tree/feature/cnn_lidar_detection}
\end{abstract}

\section{INTRODUCTION}
Automated driving can bring significant benefits to human society. It can ameliorate accidents on the road, provide mobility to persons with reduced capacities, automate delivery services, help the elderly to safely move between places, among many others.

Deep Learning has been applied successfully to detect objects with great accuracy on point cloud. Previous work on this area, used similar techniques taken from the work on Convolutional Neural Networks for object detection on images (e.g., 3D-SSMFCNN~\cite{3DSSMFCNN}), and extended its use on point cloud projections. To name some: F-pointNet~\cite{FpointNet}, VoxelNet~\cite{VxNet}, AVOD~\cite{AVOD}, DoBEM~\cite{DoBEM}, MV3D~\cite{MV3D}, among others. However, none of these are capable to perform on real-time scenarios. These approaches will be examined further in Section \ref{Related}.

Knowing that point cloud can be treated as an image (if projecting it to a 2D plane) and exploit the powerful deep learning techniques from CNNs to detect objects on it, we propose a 3D LiDAR-based multi-class object detection network (LMNet). Our network aims to achieve real-time performance, so it can be applied to automated driving systems. In contrast to other approaches, such as MV3D, which uses multiple inputs: Top and Frontal projections, and RGB data taken from a camera. Ours adopts a single-stage strategy from a single point cloud. LMNet also employs a custom designed layer for dilated convolutions. This extends the perception regions, conducts pixel-wise segmentation and fine-tunes 3D bounding box prediction. Furthermore, combining it with a dropout strategy~\cite{Dropout} and data augmentation, our approach shows significant improvements in runtime and accuracy. The network can perform oriented 3D box regression, to predict location, size and orientation of objects in 3D space.

\textbf{The main contributions of this work are}: 
\begin{itemize}
	\item To design a CNN capable to achieve real-time like 3D multi-class object detection using only point cloud data on real-time on CPU.
    \item To implement, test our work in a real vehicle.
    \item To enable multi-class object detection in the same network (Vehicle, Pedestrian, Bike or Bicycle, as defined in the KITTI dataset).
    \item To open-source the pre-trained models and inference code.
\end{itemize}

As for the evaluation of the 3D Object detection, we use the KITTI dataset~\cite{Kitti} evaluation server to submit and get a fair comparison. Experimental results suggest that LMNet can detect multi-class objects with certain accuracy for each category while achieving up to $50$ FPS on GPU. Performing better than those of DoBEM and 3D-SSMFCNN.

The structure of this paper is as follows: Section~\ref{Related} introduces the state-of-the-art architectures. Section~\ref{Proposed} details the input encoding and the LMNet. {Section~\ref{experiment} describes the network outputs, compares with other state-of-the-art architectures, and shows the tests on a real vehicle. Finally, we summarize and conclude in the Section~\ref{conclusions}.

\section{Related Work}\label{Related}
Previous research has focused primarily on improving the accuracy of the object detection. Thanks to this, great progress has been made in this field. However, these approaches leave performance out of the scope. In this section, we present our findings on the current literature and analyze how to improve performance.

\begin{table*}[h]
\caption{Implemented dilated layers}
\label{dilated_conv}
\begin{center}
\begin{tabular}{|c|c|c|c|c|c|c|c|c|} 
\hline
Layer & 1 & 2 & 3 & 4 & 5 & 6 & 7 & 8 \\ \hline
Kernel size & $3\times3$ & $3\times3$ & $3\times3$ & $3\times3$ & $3\times3$ & $3\times3$ & $3\times3$ & $1\times1$\\ \hline
Dilation & 1 & 1 & 2 & 4 & 8 & 16 & 32 & N/A \\ \hline
Receptive filed & $3\times3$ & $3\times3$ & $7\times7$ & $15\times15$ & $31\times31$ & $63\times63$ & $127\times127$ & $127\times127$\\ \hline
Feature channels & 128 & 128 & 128 & 128 & 128 & 128 & 128 & 64\\ \hline
Activation function & Relu & Relu & Relu & Relu & Relu & Relu & Relu & Relu\\ \hline
\end{tabular}
\end{center}
\vspace{-8mm}
\end{table*}

While surveying current work in this field, we found out that object detection on LiDAR data can be classified into three main different methods to handle the point cloud: a$)$ direct manipulation of raw 3D coordinates, b$)$ point cloud projection and application of Fully Convolutional Networks, and c$)$ Augmentation of the perceptive fields from previous approach using dilated convolutions.

\subsection{Manipulation of 3D coordinates}
Within this class, we can find noteworthy mentions: PointNet~\cite{PointNet}, PointNet++~\cite{PointNet++} , F-PointNet~\cite{FpointNet}, VoxelNet, VeloFCN~\cite{VeloFCN}, and MV3D. Being PointNet the pioneer on this group and can be further classified into three sub-groups.

The first group is composed of PointNet and its variants. These methods extract point-wise features from raw point cloud data. Pointnet++ was developed on top of PointNet to handle object scaling. Both PointNet and PointNet++ have shown to work reliably well in indoor environments. More recently, F-Pointnet was developed to enable road object detection in urban driving environments. This method relies on an independent image-based object detector to generate high-quality object proposals. The point cloud within the proposal boxes is extracted and fed onto the point-wise based object detector, this helps to improve detection results on ideal scenarios. However, F-PointNet uses two different input sources, cannot be trained in an end-to-end manner, requiring the image based proposal generator to be independently trained. Additionally, this approach showed poor performance when image data from low-light environments were used to get the proposals, making it difficult to deploy on real application scenarios.

The second group of methods is Voxel based. VoxelNet~\cite{VxNet} divides the point cloud scene into fixed size 3D Voxel grids. A noticeable feature is that VoxelNet directly extracts the features from the raw point cloud in the 3D voxel grid. This method scored remarkably well in the KITTI benchmark.

Finally, the last group of popular approaches is 2D based methods. VeloFCN was the first one to project the point cloud to an image plane coordinate system. Exploiting the lessons learned from the image based CNN methods, it trained the network using the well-known VGG16~\cite{vgg16} architecture from Oxford. On top this feature layers, it added a second branch to the network to regress the bounding box locations. MV3D, an extended version of VeloFCN, introduced multi-view representations for point cloud by incorporating features from bird and frontal-views.

Generally speaking, 2D based methods have shown to be faster than those that work directly on 3D space. With speed in mind, and due to the constraint we previously set of using nothing more than LiDAR data, we decided to implement LMNet as a 2D based method. This not only allows LMNet to perform faster, but it also enables it to work on low-light scenarios, since it is not using RGB data.

\subsection{Fully Convolutional Networks}

FCN have demonstrated state-of-the-art results in several semantic segmentation benchmarks (e.g., PASCAL VOC, MSCOCO, etc.), and object detection benchmarks (e.g., KITTI, etc.). The key idea of many of these methods is to use feature maps from pre-trained networks (e.g., VGGNet) to form a feature extractor. SegNet~\cite{SegNet-unpooling} initially proposed an encoder-decoder architecture for semantic segmentation. During the encoding stage, the feature map is down-sampled and later up-sampled using an unpooling layer~\cite{SegNet-unpooling}. DeepLabV1~\cite{DeepLabNet} increases the receptive field at each layer using dilated convolution filter.

In this work, the 3D point cloud is represented as a plane that extracts the shape and depth information. Using this procedure, an FCN can be trained from scratch using the KITTI dataset data or its derivatives (e.g., data augmentation). This also enables us to design a custom network, and be more flexible while addressing the segmentation problem at hand. For instance, we can integrate a similar encoder-decoder technique as the one described in SegNet. More specifically, LMNet is designed to have larger perceptive fields, quickly process larger resolution feature maps, and simultaneously validate the segmentation accuracy~\cite{Wu2016}.

\subsection{Dilated convolution}

Traditional convolution filter limits the perceptive field to uniform kernel sizes (e.g., $3 \times 3$, $5 \times 5$, etc). An efficient approach to expanding the receptive field, while keeping the number of parameters and layers small, is to employ the dilated convolution. This technique enables an exponential expansion of the receptive field while maintaining resolution~\cite{LoDNN,Yu2015} (e.g., the feature map size can be as big same as the input size). For this to work effectively, it is important to restrict the number of layers to reduce the FCN's memory requirements. This is especially true when working with higher resolution feature maps. Dilated convolutions are widely used in semantic segmentation on images. To the best of our knowledge, LMNet is the first network that uses dilated convolution filter on point cloud data to detects objects.

Table~\ref{dilated_conv} shows the implemented dilated layers in LMNet. The dilated convolution layer is larger than the input features maps, which have a size of $64\times 512$ pixels. This allows the FCN to access a larger context window to refer road objects. For each dilated convolution layer, a dropout layer is added between convolution and ReLU layers to regularize the network and avoid over-fitting.

\subsection{Comparison}\label{comparison}

\begin{table}[t]
\caption{Summary of network architectures with input Data, Detection type, Source code availability and Inference time.}
\label{basic_info}
\begin{center}
\begin{tabular}{|c|c|c|c|c|c|} 
\hline
Network 					& Input Data & Class & Code & Inference \\ \hline
F-pointNet~\cite{FpointNet} & Image and Lidar & multi & Yes & 170ms\\ \hline
VoxelNet~\cite{VxNet} 		& Lidar & multi & N/A & 30ms\\ \hline
AVOD~\cite{AVOD} 			& Image and Lidar & multi & Yes & 80ms\\ \hline
MV3D~\cite{MV3D_code} 		& Image and Lidar & car & Yes & 350ms\\ \hline
DoBEM~\cite{DoBEM} 			& Image and Lidar 	& car & N/A & 600ms \\ \hline
3D-SSMFCNN~\cite{3DSSMFCNN} & Image 	& car & Yes & 100ms\\ \hline
\end{tabular}
\end{center}
\vspace{-5mm}
\end{table}

\begin{figure}[t]
\centering
\includegraphics[width=0.45\textwidth]{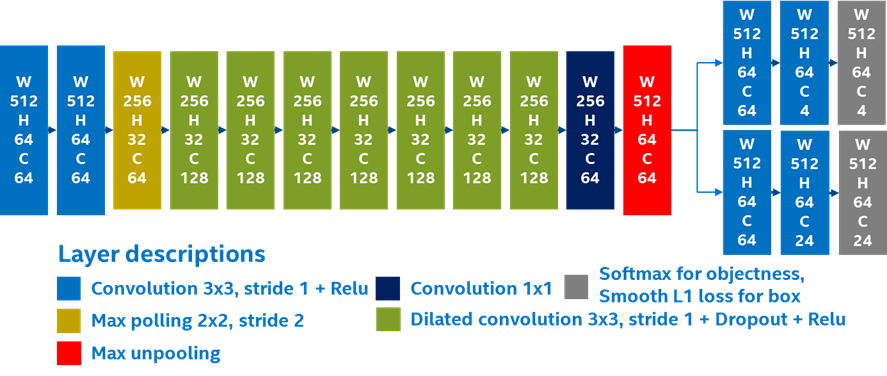}
\caption{LMNet architecture}
\label{architecture}
\vspace{-5mm}
\end{figure}

As this work aims to design a fast 3D multi-class object detection network, we mainly compare with other previously published 3D object detection networks: F-pointNet, VoxelNet, AVOD, MV3D, DoBEM, and 3D-SSMFCNN.

To have a basic overview of the previously mentioned networks, Table~\ref{basic_info} lists the items: (a) Data, which is what data feed in to the network; (b) type, which is detection availability (i.e. multi-class, car only); (c) Code, which is code availability, and; (d) Inference, which is the time taken to forward the input data and generate outputs.

Although, some of state-of-the-art network source code (viz., F-pointNet, AVOD) are available, the models reported to Kitti are not available. We can only infer that by the type of data being fed, the outlined accuracy, and the manually reported inference time. Additionally, none of them are capable to be processed in real-time (more than 30 FPS)~\cite{real-time}. A third party, Boston Didi team, implemented the MV3D model~\cite{MV3D_code}. This method can only perform single class detection and would not be suitable to perform a comparison. Furthermore, the reported inference performance is less than 3 FPS. For the previous reasons, we firmly believe that open sourcing a fast object detection method and its corresponding inference code can greatly contribute to the development of faster and highly accurate detectors in the automated driving community.

\section{Proposed Architecture}\label{Proposed}
The proposed network architecture takes as input five representations of the frontal-view projection of a 3D point cloud. These five input maps help the network to keep 3D information at hand. The architecture outputs an objectness map and 3D bounding box offset values calculated directly from the frontal-view map. The objectness map contains the class confidence values for each of the projected 3D points. The box candidates calculated from the offset values are filtered using a custom euclidean distance based non-maximum suppression procedure. A diagram of the proposed network architecture can be visualized in Fig.~\ref{architecture}.

Albeit frontal-view representations~\cite{3DVelo} have less information than bird-view~\cite{LoDNN,MV3D} representations or raw-point~\cite{PointNet,PointNet++} data. We can expect less computational cost and certain detection accuracy~\cite{Vote3D}. 

\subsection{Frontal-view representation\label{FV_representation}}

To obtain a sparse 2D point map, we employ a cylindrical projection~\cite{3DVelo}. Given a 3D point $p = (x, y, z)$. Its corresponding coordinates in the frontal-view map $p_{f} = (r, c)$ can be calculated as follows:

\begin{align}
c &= \left \lfloor atan2(y,x) / \delta \theta \right \rfloor ,\\
r &= \left \lfloor atan2(z, \sqrt[]{x^2 + y^2}) / \delta \phi \right \rfloor .
\end{align}

\noindent where $\delta \theta$ and $\delta \phi$ are the horizontal and vertical resolutions (e.g., $0.32$ and $0.4$ while $0.08$, $0.4$ radian in Velodyne HDL-64E~\cite{HDL64E}) of the LiDAR sensor, respectively. 

Using this projection, five feature channels are generated:

\begin{enumerate}
    \item Reflection, which is the normalized value of the reflectivity as returned by the laser for each point.
    \item Range, which is the distance on the XY plane (or the ground) of the sensor coordinate system. It can be calculated as $\sqrt{x^2 + y^2}$.
    \item The distance to the front of the car. Equivalent to the x coordinate on the LiDAR frame. 
    \item Side, the y coordinate value on the sensor coordinate system. Positive values represent a distance to the left, while negative ones depict points located to the right of the vehicle.
    \item Height, as measured from the sensor location. Equal to the z coordinate as shown in Fig.~\ref{FVs}.
\end{enumerate}

\begin{figure}[t]
\centering
\includegraphics[width=0.35\textwidth]{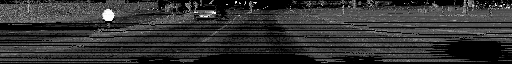}\\
(a) Reflection\\
\includegraphics[width=0.35\textwidth]{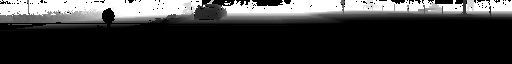}\\
(b) Range\\
\includegraphics[width=0.35\textwidth]{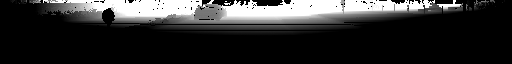}\\
(c) Forward\\
\includegraphics[width=0.35\textwidth]{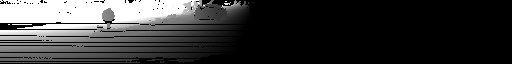}\\
(d) Side \\
\includegraphics[width=0.35\textwidth]{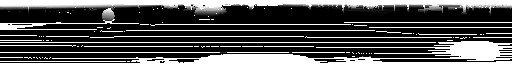}\\
(e) Height\\
\caption{Encoded input point cloud\label{FVs}}
\vspace{-7mm}
\end{figure}

\subsection{Bounding Box Encoding}

As for the bounding boxes, we follow the encoding approach from ~\cite{3DVelo}, which considers the offset of corner points and its rotation matrix. There are two reasons on why to use this approach as shown in Ref~\cite{3DVelo}: (A) A faster CNN convergence due to the reduced offset distribution leading to a smaller regression search space, and (B) Enable rotation invariance. We briefly describe the encoding in this section.

Assuming a LiDAR point $p = (x, y, z) \in P$, an object point and a background point can be represented as $p \in O$, $p \in O^{c}$, respectively. The box encoding considers the points forming the objects, e.g., $p \in O$. The observation angles (e.g., azimuth and elevation) are calculated as follows:

\begin{align}
\theta &= atan2(y,x),\\
\phi &= atan2(z, \sqrt{x^2 + y^2}).
\end{align}

\noindent Therefore, the rotation matrix $R$ can be defined as follows,

\begin{equation}
R = R_{z}(\theta)R_{y}(\phi).
\end{equation}

\noindent where $R_{z}(\theta)$ and $R_{y}(\phi)$ are the rotation functions around the z and y axes, respectively. Then, the $i$-th bounding box corner $c_{p,i} = (x_{c,i}, y_{c,i}, z_{c,i})$ can be encoded as:

\begin{equation}\label{offset}
c'_{p,i} = R^{T}(c_{p,i} - p).
\end{equation}

\noindent Our proposed architecture regress $c'_{p}$ during training. The eight corners of a bounding box are concatenated in a 24-d vector as,

\begin{equation}
b'_{p} = (c'^{T}_{p,1},c'^{T}_{p,2},c'^{T}_{p,3}, ... , c'^{T}_{p,8})^{T}.
\end{equation}

\noindent Thus, the bounding box output map has 24 channels with the same resolution as the input one.

\subsection{Proposed Architecture}
The proposed CNN architecture is similar to LoDNN~\cite{LoDNN}. As illustrated in Fig.~\ref{architecture}, the CNN feature map is processed by two $3 \times 3$ convolution, eight dilated convolution and followed by $3 \times 3$ and $1 \times 1$ convolution layers. The trunk splits at the max-pooling layer into the objectness classification branch and the bounding box's corners offset regression branch. We use $(d)conv(c_{in},c_{out},k)$ to represent a 2-dimensional convolution/dilated-convolution operator where $c_{in}$ is the number of input channel, and $c_{out}$ is the number of output channels, $k$ represent the kernel size, respectively.

A five-channel map of size of $64 \times 512 \times 5$, as described in Section~\ref{FVs}, are fed into an encoder with two convolution layers, $conv(5,64,3)$ and $conv(64,64,3)$. These are followed by a max-pooling layer to the output of the encoder, this helps to reduce FCN's memory requirements as previously mentioned.

The encoder section is succeeded by seven dilated-convolutions~\cite{Yu2015} \
(see dilation parameters in Table~\ref{dilated_conv}
) and a convolution, e.g., $dconv_{1}(64,128,3)$; $dconv_{2-7}(128,128,3)$; and $conv(128,64,1)$}, with a dropout layer and a rectified linear unit (ReLU) activation applied to the pooled feature map, enabling the layer to extract multi-scale contextual information.

Following the context module, the main trunk bifurcates to create the objectness and corners branches. Both of them have a decoder that up-samples the feature map, from the dilated convolutions to the same size as the input. This is achieved with the help of a max-unpooling layer~\cite{SegNet-unpooling}, followed by two convolution layers: $conv(64,64,3)$ with ReLU, and; $conv(64,4,3)$ with ReLU for objectness branch, or $conv(64,24,3)$ for corners branch.

The objectness branch outputs an object confidence map, while the corners-branch generates the corner-points offsets. Softmax loss is employed to calculate the objectness loss between the confidence map and the encoded objectness map. Finally, Smooth $l_{1}$~\cite{Girshick2015} is utilized for the corner offset regression loss between the corners offset and ground truth corners offset.

To train a multi-task network, a re-weighting approach is employed as explained in Ref~\cite{3DVelo} to balance the objective of learning both objectness map and bounding box's offset values. The multi-task
function $L$ can be defined as follows:

\begin{equation}
L = \sum_{p \in P} w_{obj}(p) L_{obj}(p) + \sum_{p \in O} w_{cor}(p) L_{cor}(p),
\end{equation}

\noindent where $L_{obj}$ and $L_{cor}$ denote the softmax loss and regression loss of point $p$, respectively, while $w_{obj}$ and $w_{cor}$ are point-wise weights, e.g., objectness weight and corners weight, calculated by the following equations:

\begin{equation}
w_{cor}(p) = \left\{\begin{matrix}
\bar{s}[\kappa_{p}]/s(p) & p \in O,\\ 
1 & p \in O^{c},
\end{matrix}\right.
\end{equation}

\begin{equation}
w_{bac}(p) = \left\{\begin{matrix}
m|O|/|O^{c}| & p \in O^{c},\\ 
1 & p \in O,
\end{matrix}\right.
\end{equation}

\begin{equation}
w_{obj}(p) = w_{bac}(p) w_{cor}(p),
\end{equation}

\noindent where $\bar{s}[x]$ is the average shape size of the class $x \in \{\text{car, pedestrian, cyclist}\}$, $\kappa_{p}$ denotes point $p$ class, $s(p)$ represents the size of an object which the point $p$ belongs to, $|O|$ denotes the number of points on all objects, $|O^{c}|$ denotes the number of points on the background. In a point cloud scene, most of the points are corresponding to the background. A constant weight, $m$, is introduced to balances the softmax losses between the foreground and the background object. Empirically, $m$ is set to 4.

\subsection{Training Phase}

The KITTI dataset only has annotations for objects in the frontal-view on the camera image. Thus, we limit point cloud range in $[0, 70]\times [-40, 40]\times [-2, 2]$ meters, and ignore the points in out of image boundaries after projection (see Section~\ref{FV_representation}). Since KITTI uses the Velodyne HDL64E, we obtain a $64\times 512$ map for the frontal-view maps.

The proposed architecture is implemented using Caffe~\cite{caffe}, while adding the custom unpooling layers implementations. The network is trained in an end-to-end manner using stochastic gradient decent (SGD) algorithm with a learning rate of $10^{-6}$ for 200 epoch for our dataset. The batch size is set to 4.

\subsection{Data augmentation}

The number of point cloud samples in the KITTI training dataset is $7481$, which is considerably less than other image datasets (e.g., ILSVRC~\cite{ILSVRC15}, has $456567$ images for training). Therefore, data augmentation technique is applied to avoid overfitting, and to improve the generalization of our model. Each point cloud cluster that is corresponding to an object class was randomly rotating at LiDAR z-axis $[-15^{\circ}, 15^{\circ}]$. Using the proposed data augmentation, the training set dramatically increased to more than 20000  data samples.

\subsection{Testing Phase}

The objectness map includes a non-object class and the corner map may output many corner offsets. Consider the corresponding corner points \{$c_{p,i}|i\in 1, ... 8\}$ by the inverse transform of Eq.~\ref{offset}. Each bounding box candidates can be denoted by $b_{p} = (c^{T}_{p,1},c^{T}_{p,2},c^{T}_{p,3}, ... ,c^{T}_{p,8})^{T}$. The sets of all box candidates is $B = \{b_{p}|p \in obj\}$. 

To reduce bounding boxes redundancy, we apply a modified non-maximum suppression (NMS) algorithm, which selects the bounding box based on the euclidean distance between the front-top-left and rear-bottom-right corner points of the box candidates. Each box $b_{p}$ is scored by counting its neighbor bounding boxes in B within distance $\delta_{car}, \delta_{pedestrian}, \delta_{cyclist}$,  denoted as $\#\{||c_{pi,1} - c_{pj,1}|| + ||c_{pi,8} - c_{pj,8}|| < \delta_{class}\}$. Then, the bounding boxes B are sorted by descending score. Afterwards, the bounding box candidates whose score is less than five are discarded as outliers. Picking up a box who has the highest score in B, the euclidean distances between the box and the rest are calculated. Finally, the boxes whose distance are less than the predefined threshold values are removed. The empirically obtained thresholds for each class are: (a) $T_{car} = 0.7m$; (b) $T_{pedestrian}=0.3m$, and; (c) $T_{cyclist}=0.3m$. 

\section{Experiments\label{experiment}}

The evaluation results of LMNet on the challenging KITTI object detection benchmark~\cite{Kitti} are shown in this section. The benchmark provides image and point cloud data, $7481$ sets for training and $7518$ sets for testing. LMNet is evaluated on KITTI's 2D, 3D and bird view object detection benchmark using online evaluation server.

\begin{table}[t]
\caption{Performance board}
\label{performance_board}
\begin{center}
\vspace{-5mm}
\begin{tabular}{|c|c|c|c|c|c|} 
\hline
Newtwork 					& Accelerator & Inference & Car & Ped. & Cyc.\\ \hline
F-PointNet~\cite{FpointNet} & GTX 1080 	& 88ms & \bf{70.39} & \bf{44.89} & \bf{56.77} \\ \hline
VoxelNet (Lidar)~\cite{VxNet} 	& Titan X 	& 30ms & 65.11 & 33.69 & 48.36 \\ \hline
AVOD~\cite{AVOD} 			& Titan Xp 	& 80ms & 65.78 & 31.51 & 44.90 \\ \hline
MV3D~\cite{MV3D} 			& Titan X 	& 360ms & 63.35 & N/A & N/A \\ \hline
MV3D (Lidar)~\cite{MV3D} 	& Titan X 	& 240ms & 52.73 & N/A & N/A \\ \hline
DoBEM~\cite{DoBEM} 			& Titan X 	& 600ms & 6.95 & N/A & N/A \\ \hline
3D-SSMFCNN~\cite{3DSSMFCNN} & Titan X 	& 100ms & 2.28 & N/A & N/A \\ \hline
Proposed 					& GTX 1080 & \bf{20ms} & 15.24 & 11.46 & 3.23 \\ \hline
\end{tabular}
\end{center}
\vspace{-9mm}
\end{table}

\subsection{Inference Time}
One of the major issue for the application of a 3D object detection network is inference time. We compare the inference time taken by LMNet and some representative state-of-art networks. Table~\ref{performance_board} shows inference time on CUDA enabled GPUs of LMNet and other representatives methods. LMNet achieved the fastest inference at $20$ms on a GTX $1080$. Further tests on a Titan Xp observed inference time of $12$ms, and $6.8$ms ($147$ FPS) for Tesla P40. LMNet is the fastest in the wild and enables real-time ($30$ FPS or better)~\cite{real-time} detection.

\subsection{Towards Real-Time Detection on General Purpose CPU}
To test the inference time on CPU, LMNet is using Intel Caffe~\cite{IntCaffe}, which is optimized for Intel architecture CPUs through Intel's math kernel library (MKL). For the execution measurements, we include not only the inference time but also the time required to pre-process, project and generate the input data from the point cloud, as well as the post-processing time used to generate the objectness map and bounding box coordinates. Table~\ref{runtime_info_cpu} shows that LMNet can achieve more than 10 FPS on an Intel CPU i5-6600K (4 cores, $3.5$ GHz) and 20 FPS on Xeon E5-2698 v4 (20 cores, $2.2$ GHz). Given that the Lidar scanning frequency is 10 Hz, the 10 FPS achieved by LMNet is considered as real-time. These results are promising for automated driving systems that only use general purpose CPU. LMNet could be deployed to edge devices. 

\begin{table}[t]
\caption{Inference time on CPU}
\label{runtime_info_cpu}
\begin{center}
\begin{tabular}{|c|c|c|} 
\hline
Caffe version 			& i5-6600K 	& E5-2698 v4 \\ \hline
Caffe with OpenBlas		& 351ms 	& 280ms\\ \hline
Intel Caffe with MKL2017 & 99ms 		& 47ms\\ \hline
\end{tabular}
\end{center}
\vspace{-4mm}
\end{table}

\subsection{Project Map Segmentation}
\begin{figure}[t]
\centering
\includegraphics[width=0.4\textwidth]{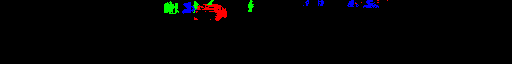}\vspace{1mm} \\
(a) Ground truth label
\includegraphics[width=0.4\textwidth]{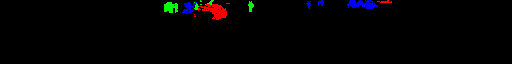} \\
(b) Segmented map
\caption{Ground truth label and segmented map}
\label{segmentation}
\vspace{-6mm}
\end{figure}

Based on our observations, the trained model at $200$ epochs is used for performance evaluation. The obtained segmented map and its corresponding ground truth label from the validation set can be appreciated in Fig.~\ref{segmentation}. The segmented object confidence map is very similar to the ground truth. Experiment result shows that LMNet can accurately classify points by its class.

\subsection{Object Detection}
The evaluation of LMNet was carried out on the KITTI dataset evaluation server. Particularly, multi-class object detection on 2D, 3D and BV categories. Table.~\ref{performance_board} shows the average precision on 3D object detection in the moderate setting. LMNet performed a certain 3D detection accuracies, $15.24\%$ for car, $11.46\%$ for pedestrian and $3.23\%$ for cyclist, respectively. The state-of-art networks achieve more than $50\%$ accuracy for car, $20\%$ for the pedestrian and $29\%$. MV3D achieved $52\%$ accuracy for car objects. However, those networks are either not executed in real-time, only single class, or not open-sourced as mentioned in the Section~\ref{comparison}. To our best knowledge, LMNet is the fastest multi-class object detection network using data only from LiDAR, with models and code open sourced.

\subsection{Test on real-world data}
\begin{figure}[t]
\centering
\includegraphics[width=0.35\textwidth]{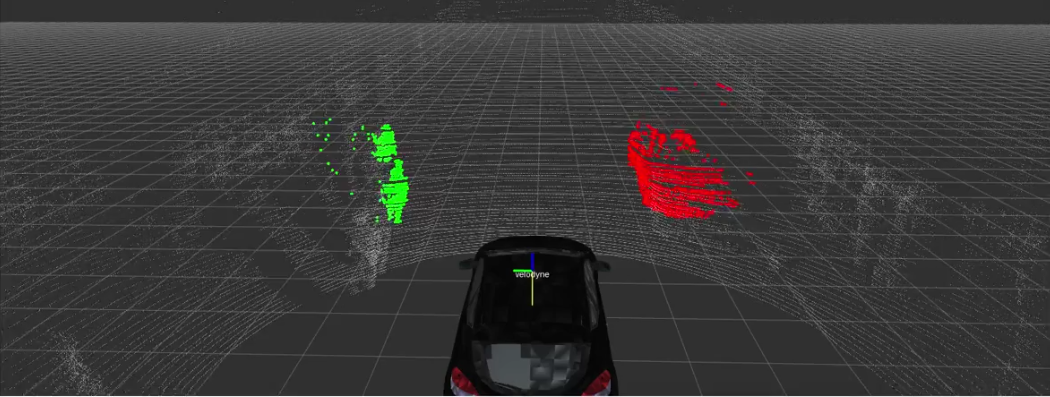}
\caption{3D point cloud classified by LMNet from a Velodyne HDL-64E }
\label{ros_pointcloud_classified}
\vspace{-6mm }
\end{figure}

We implemented LMNet as an Autoware~\cite{Autoware} module. Autoware is an autonomous driving framework for urban roads. It includes sensing, localization, fusion and perception modules. We decided to use this framework due to the easiness of installation, its compatibility with ROS~\cite{ROS} and because it is open-source, allowing us to focus only on the implementation of our method. The sensing module allowed us to connect with a Velodyne HDL-64E LiDAR, the same model as the one used in the KITTI dataset. With the help of the ROS, PCL and OpenCV libraries, we projected the sensor point cloud, constructed the five-channel input map as described in ~\ref{FV_representation}, fed them to our Caffe fork and obtained both the output maps. Fig.~\ref{ros_pointcloud_classified} shows the classified 3D point cloud.

\section{CONCLUSIONS\label{conclusions}}
This paper introduces LMNet, a multi-class, real-time network architecture for 3D object detection on point cloud. Experimental results show that it can achieve real-time performance on consumer grade CPUs. The predicted bounding boxes are evaluated on KITTI evaluation server. Although the accuracy is on par with other state-of-the-art architectures, LMNet is significantly faster and able to detect multi-class road objects. The implementation and pre-trained models are open-sourced, so anyone can easily test our claims. Training code is also planned to be open sourced. It is important to note that all the evaluations are using the bounding box location, as defined by the KITTI dataset. However, this does not correctly reflect the classifier accuracy at a point-wise level. As for future work, we intend to fine-tune the bounding box non-maximum suppression algorithm to perform better on the KITTI evaluation; Implement the network on low-power platforms such as the Intel Movidius's Myriad~\cite{myriad} and Atom, to test its capability on a wider range of IoT or embedded solutions.



\balance
\bibliographystyle{IEEEtran}
\bibliography{references}

\end{document}